\documentclass{article}
\usepackage{arxiv}

\usepackage{graphicx}
\usepackage[export]{adjustbox}
\usepackage{amssymb}
\usepackage{amsmath}
\usepackage[dvipsnames]{xcolor}
\definecolor{lightgray}{gray}{0.8}
\definecolor{lightlightgray}{gray}{0.9}

\usepackage{tikz}

\newtheorem{mydefi}{Definition}
\usepackage{hyperref}

\begin{document}
\suppressfloats 

\tikzset{fontscale/.style = {font=\relsize{#1}}}

\title{Analysis and Prediction of Deforming 3D Shapes using Oriented Bounding Boxes and LSTM Autoencoders} 
\author{Sara Hahner \\
        Fraunhofer Center for Machine Learning and SCAI \\
        {\tt\small sara.hahner@scai.fraunhofer.de}\\
	    \And Rodrigo Iza-Teran \\
	    Fraunhofer Center for Machine Learning and SCAI\\
		{\tt\small rodrigo.iza-teran@scai.fraunhofer.de}\\
		\And  Jochen Garcke \\
	    Fraunhofer Center for Machine Learning and SCAI\\
	    University of Bonn\\
		{\tt\small jochen.garcke@scai.fraunhofer.de}\\
}

\newcommand{\thedate}{August 14, 2020} 

\maketitle             
\begin{abstract}
For sequences of complex 3D shapes in time we present a general approach to detect patterns for their analysis and to predict the deformation by making use of structural components of the complex shape. 
We incorporate long short-term memory (LSTM) layers into an autoencoder to create low dimensional representations that allow the detection of patterns in the data and additionally detect the temporal dynamics in the deformation behavior. This is achieved with two decoders, one for reconstruction and one for prediction of future time steps of the sequence. 
In a preprocessing step the components of the studied object are converted to oriented bounding boxes which capture the impact of plastic deformation and allow reducing the dimensionality of the data describing the structure. The architecture is tested on the results of 196 car crash simulations of a model with 133 different components, where material properties are varied. In the latent representation we can detect patterns in the plastic deformation for the different components. The predicted bounding boxes give an estimate of the final simulation result and their quality is improved in comparison to different baselines.

\end{abstract}

\section{Introduction} 

\begin{figure}[t]
\center
\begin{minipage}[c]{0.12\textwidth}
\center
\footnotesize{Mode A:}
\end{minipage}
\begin{minipage}[c]{0.3\textwidth}
\center
\includegraphics[width=\textwidth,trim={32cm 5cm 27.5cm 22cm},clip]{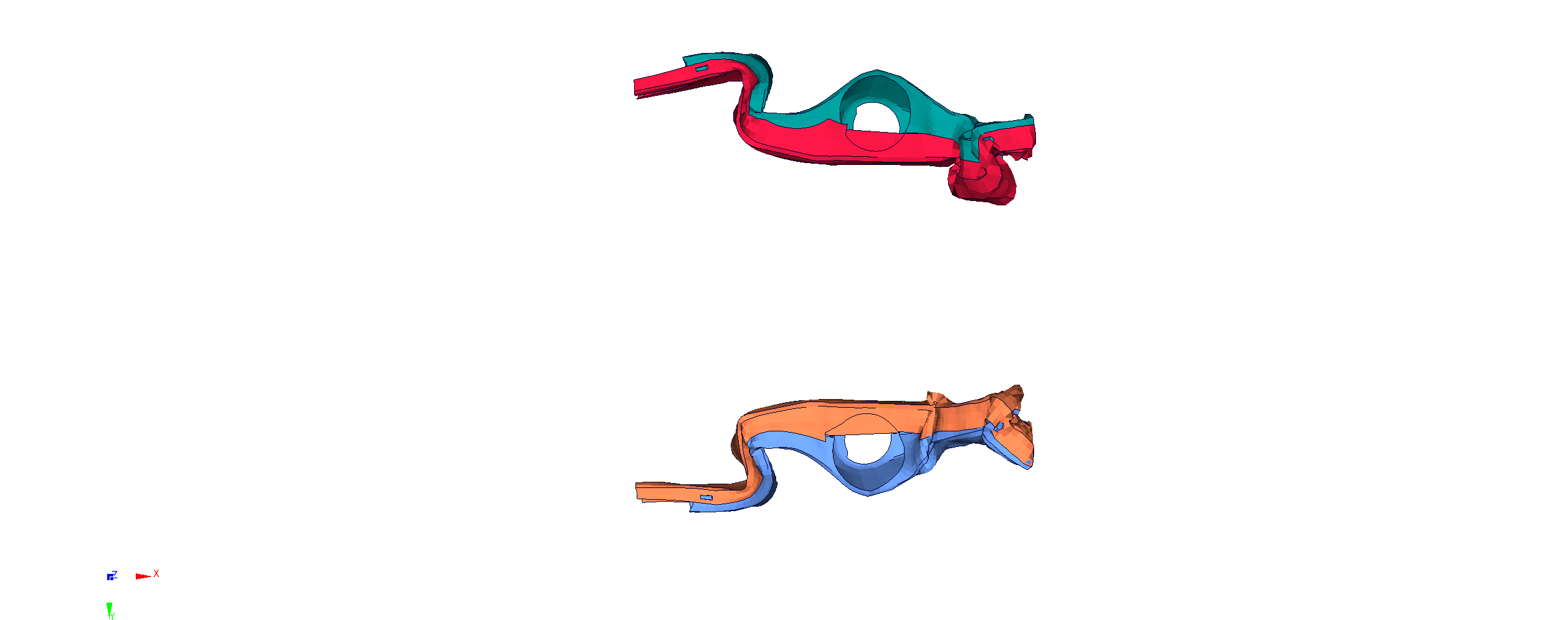}
\end{minipage}
\hspace{2ex}
\begin{minipage}[c]{0.12\textwidth}
\center
\footnotesize{Mode B:}
\end{minipage}
\begin{minipage}[c]{0.3\textwidth}
\center
\includegraphics[width=\textwidth,trim={31cm 5cm 27.5cm 22cm},clip]{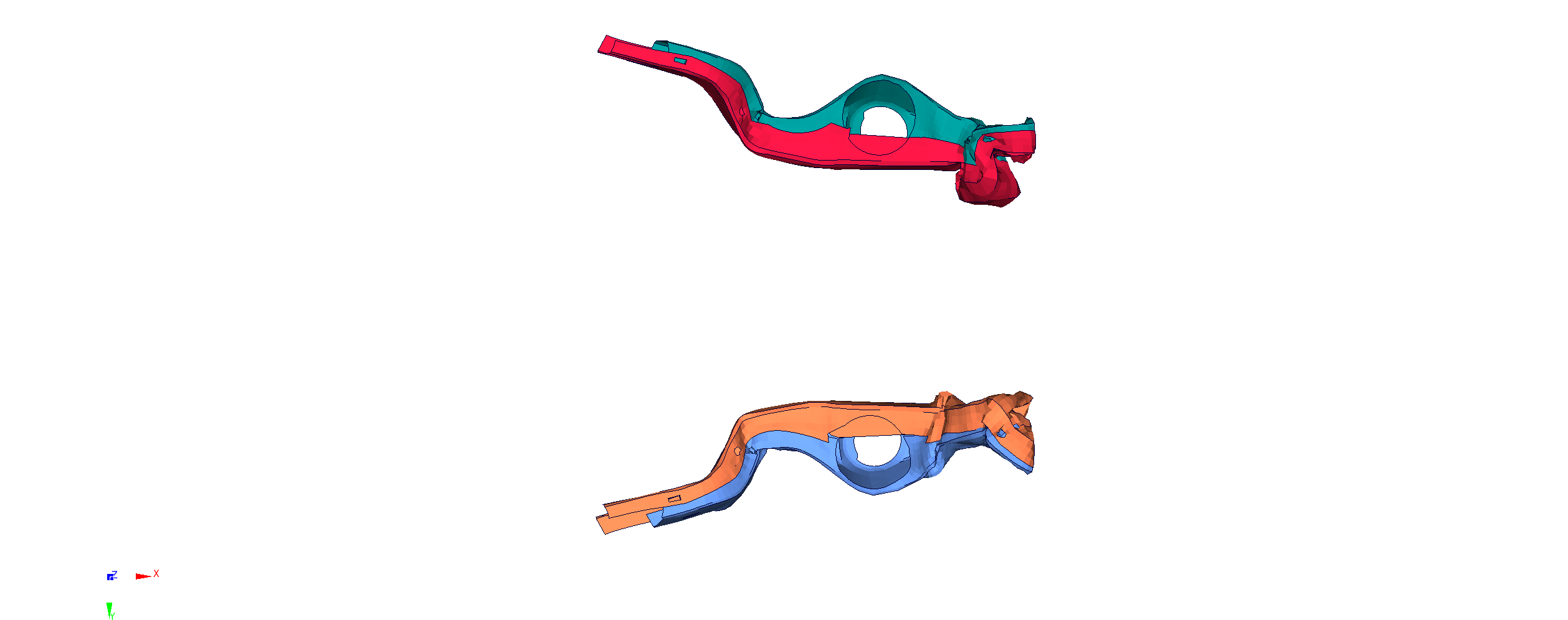}
\end{minipage}
\caption{Deformation modes of left front beams in the example car model.}
\label{snapshot2}
\end{figure}

Deforming 3D shapes that can be subdivided into components can be found in many different areas, for example crash analysis, structure, and material testing, or moving humans and animals that can be subdivided into body parts.
The data can be especially challenging since it has a temporal and a spatial dimension that have to be considered jointly.
In particular, we consider shape data from Computer Aided Engineering (CAE), where numerical simulations play a vital role in the development of products, as they enable simpler, faster, and more cost-effective investigations of systems. 
If the 3D shape is complex, the detection of patterns in the deformation, called deformation modes in CAE (figure \ref{snapshot2}), speeds up the analysis.

While deformation modes are dependent on model parameters, their behavior is accessible, albeit tediously, upon manifestation when having access to the completed simulation run. In practice this detection of deformation modes is based on error-prone supervision of a non-trivial, hand-selected subset of nodes in critical components.

In general, simulation results contain abundant features and are therefore high dimensional and 
unwieldy, which makes comparative analysis difficult. Intuitively, the sampled points describing the component are highly correlated, not only in space but also over time.   
This redundancy in the data invites further analysis via feature learning, reducing unimportant information in space as well as time and thus dimensionality. 
The low dimensional features capture distinguishing characteristics to detect the deformation modes.
As soon as the deformation of a component is attributed to a deformation mode, there is little variance regarding its further course in the simulation. This motivates the in-situ detection, that is during the simulation, of deformation modes; we can then interrupt the simulation and anticipate targeted changes to the model parameters.

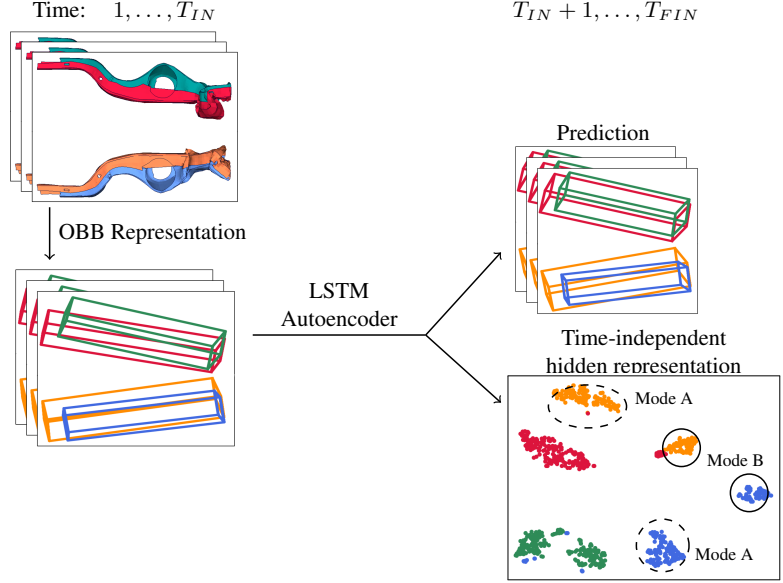
\begin{figure}[t]
\center
\begin{minipage}[c]{0.78\textwidth}
\center
\begin{tikzpicture}[line width=0.2mm, every node/.style={scale=0.95}]
\node[align=center] at (0,1.4) {\small{Time: \hspace{2ex}$1, \dots, T_{IN}$}};

\node[align=center] at (0,0) {\includegraphics[width=0.25\textwidth]{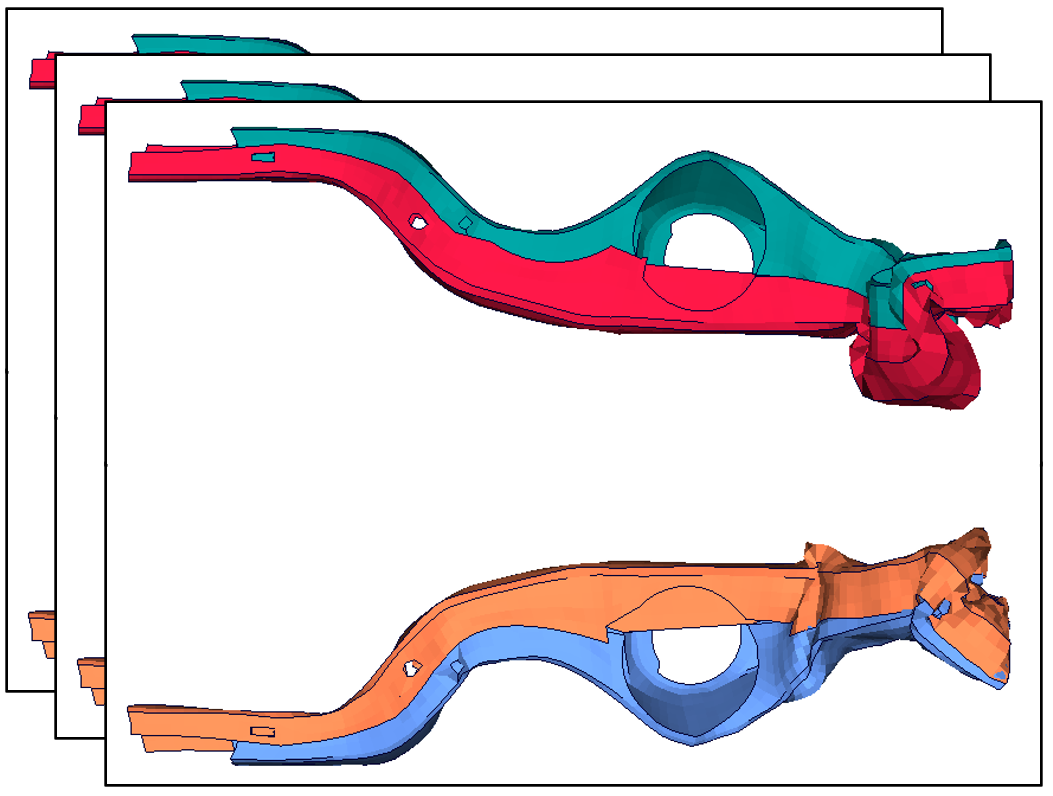}};
\draw[->] (-1,-1.2) -- node[right] {\footnotesize{OBB Representation} } (-1,-1.9);
\node[align=center] at (0,-3.2) {\includegraphics[width=0.24\textwidth]{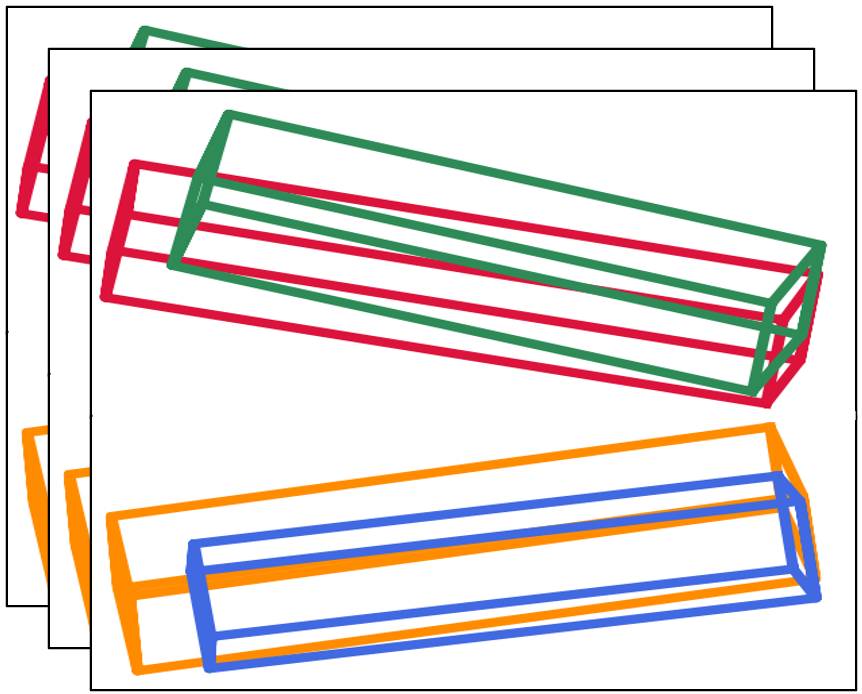}};

\draw[-]  (1.7,-2.9) -- node[above,align=center] {\small{LSTM} \\ \small{Autoencoder} } (4,-2.9);

\node[align=center] at (6.4,1.4) {\small{$T_{IN}+1, \dots, T_{FIN}$}};

\draw[->] (4,-2.9) --  (5,-1.8);
\node[align=center] at (6.4,-1.5) {\includegraphics[width=0.2\textwidth]{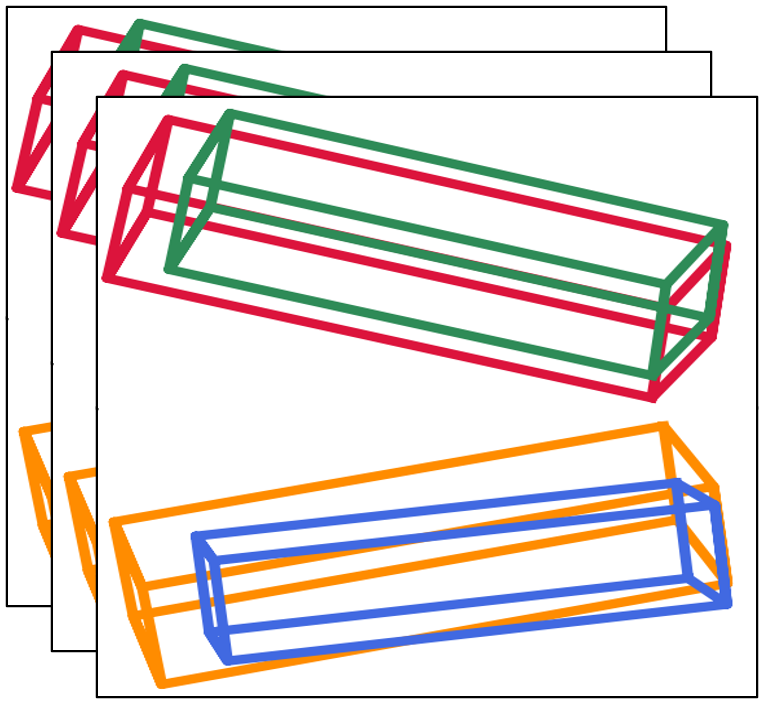}};
\node[align=center] at (6.4,-0.2) {\small{Prediction} };

\draw[->] (4,-2.9) --  (5,-3.8);
\node[align=center] at (6.9,-3.15) {\small{Time-independent} \\ \small{hidden representation} };
\setlength{\fboxrule}{0.3pt}
\setlength{\fboxsep}{-1pt}
\node[align=center] at (6.9,-4.8) {{\includegraphics[width=0.3\textwidth,trim={6.8cm 4cm 13.3cm 3.5cm},clip,
fbox]{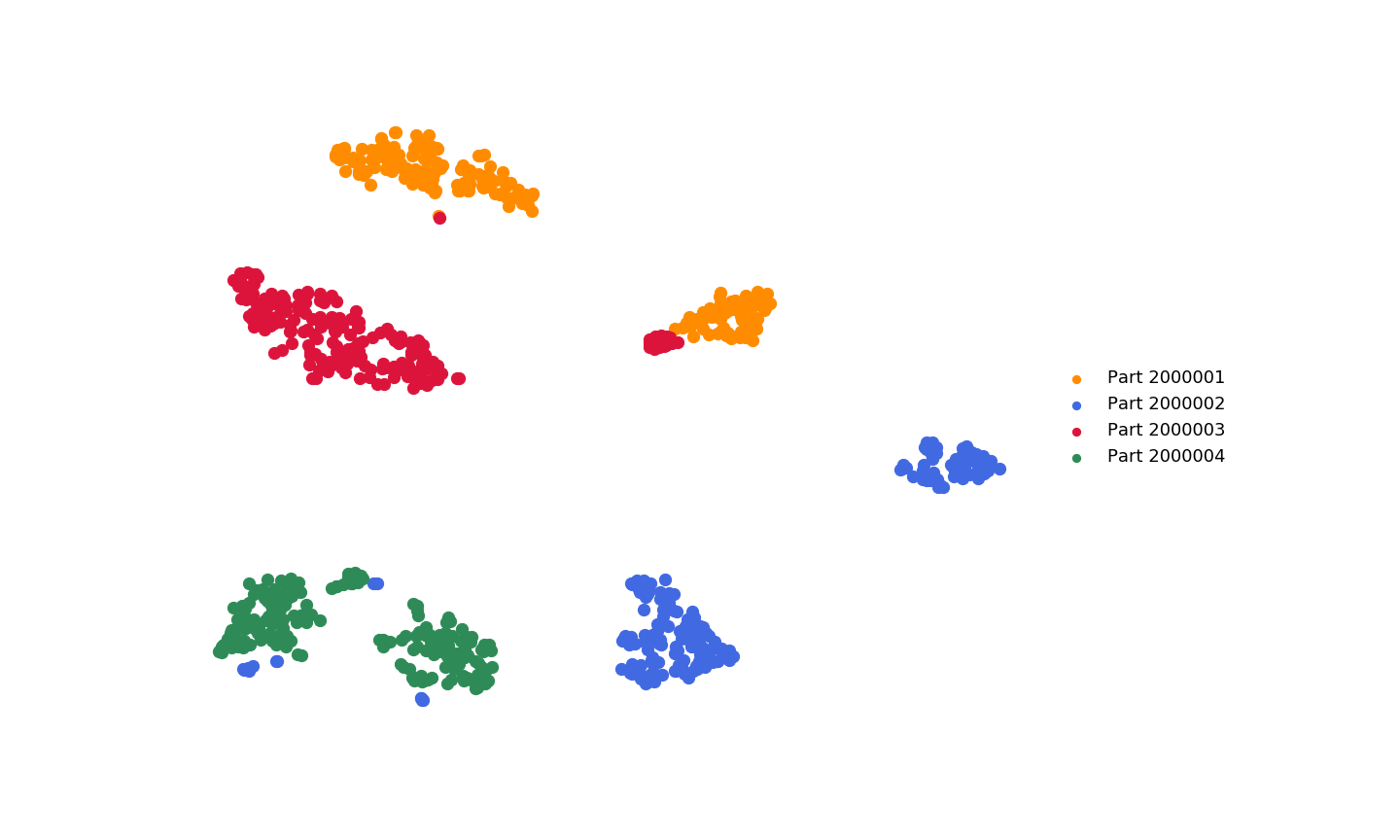}}};
\draw (7.4,-4.4) circle (0.25cm); 
\draw (8.3,-5) circle (0.25cm);
\draw[dashed] (7.15,-5.7) circle (0.35cm);
\node [rotate=0][draw,dashed,inner sep=11pt, circle,yscale=.55] at (6.15,-3.85) {};
\node at (8.12,-4.57) [font=\scriptsize] {Mode B};
\node at (7.17,-3.74) [font=\scriptsize] {Mode A};
\node at (7.97,-5.82) [font=\scriptsize] {Mode A};
\end{tikzpicture}
\end{minipage}
\caption{Analysis and Prediction of Deforming Shapes. Preprocess the first $T_{IN}$ time steps of simulation data by oriented bounding boxes, apply an LSTM autoencoder which predicts the future time steps and creates a hidden representation representing the deformation over time. Finally, the deformation modes from figure \ref{snapshot2} can be detected in the hidden representations.} \label{method}
\end{figure}

For the in-situ analysis of deforming shapes, we propose the application of an LSTM autoencoder trained on a component-wise shape representation by oriented bounding boxes instead of directly on the 3D data points (or meshes) describing the shapes.
Motivated by an architecture with two decoders that stems from unsupervised video prediction \cite{UnsupLSTM}, we choose a composite LSTM autoencoder that is specialized on handling time sequences and can both reconstruct the input and predict the future time steps of a sequence.
It creates an additional low dimensional hidden or latent representation which after successful training represents the features of the input time sequences and allows its reconstruction and prediction. In this case the features encode the distinctive characteristics of the component's deformation. Figure \ref{method} illustrates the process.

The research goals can be summarized as a) the definition of a distinguishing and simple shape representation for structural components of a complex model, b) the in-situ detection of patterns in the deformation behavior, and 
c) the in-situ prediction of structural component positions at future time steps.

Further on in section \ref{sec:sim}, we give an introduction to simulation data, which we choose as example data, its characteristics and limitations. 
In section \ref{sec:obb}, we present oriented bounding boxes, the selected shape representation for structural components, followed by the definition of the LSTM autoencoder in section \ref{sec:LSTM}. 
In section \ref{sec:rel}, we discuss related work on machine learning for simulations. 
Results on a car crash simulation are presented in section \ref{sec:exp}.

\section{Simulation Data and Their Characteristics}
\label{sec:sim}

As an example for a deforming shape over time, we consider 3D (surface) meshes, which are decomposed into structural components. 
A numerical simulation approximately computes the deformation of the underlying object on the basis of physical relations. 
For each simulation, model, material, or environment parameters might be varied as well as the geometry of components or their connections.
The output of a simulation are the positions of the mesh points at different times and point-wise values of physical quantities. 
While a large variety of deformation processes is observed per component, they result in a small number of meaningful patterns, called deformation modes.
In figure \ref{snapshot2} two distinct deformation modes of two structural components during a car crash can be observed.

Car crash simulations are a complex example. The mesh sizes for the car models can be large, the number and diversity of the input parameters is high and the behavior is nonlinear. Furthermore, the calculations need to have a high degree of precision, because the evaluation of the passenger's safety is based on the simulation outputs, i.e. plastic deformation, strain, and rupture. 
The stored simulations nowadays contain detailed information for up to two hundred time steps and more than ten million nodes. 
Due to the large computational costs of one simulation, for one model there are generally less than 500 different runs. 

From simulation run to simulation run the input parameters are modified to achieve the design goals, e.g. crash safety, weight, or performance. 
In the beginning, the goal is the broad understanding of deformation behavior by applying major changes to the model. In later design stages a fine tuning is done and tests address some issues that have not been captured in early stages.
We present an approach that analyzes the whole car as a multi-body object under dynamic deformation. Therefore, it is applied at an early design stage where the broad deformation behavior and component interaction has to be understood.

\subsubsection*{Challenges in the Analysis}
Because of the high data complexity, the relatively low number of samples, and fatal consequences of wrong results on the passenger's safety, many data analysis methods cannot be applied directly. Furthermore, a full prediction of the complex simulation results, including detailed deformation or strains, only based on the input parameters is error prone and critical for the purpose of precise parameter selections and the understanding of detailed physical behavior.  
Existing methods create a surrogate model given some input parameters, see e.g. \cite{Steffes}. Others show how deformation modes depend on input parameters by using visualizations in low dimensional representations without defining an explicit surrogate model \cite{iza2019geometrical,Thole}.

We do not want to base our analysis on input parameters, but present a more general approach that concentrates on general deformation behavior. 
By applying an in-situ analysis and an estimation of the deformation in the future time steps, we provide information that allows a faster selection of new model parameters in the development process.

Different mesh resolutions or changes in the geometry of two different models lead to a correspondence problem. The 3D surface meshes cannot be compared point-wise anymore requiring the use of shape representations of the meshes, that capture the differences between different modes of deformation (figure \ref{snapshot2}). 
However, semantic design rules apply for the car model’s components as is the case for most man-made models.
Car wheels, the seats, and beams are all of similar shape for different car models, which can be inferred once the general form and dimension are known, especially when analyzing broad characteristics in the early design stages. Because of these semantic rules, the model’s components can be approximated by their optimal bounding boxes which capture their translation, rotation, and scale. 

\section{Preprocessing by Oriented Bounding Boxes}  
\label{sec:obb}

We preprocess every structural component by determining an oriented bounding box for each time step. For that, we consider the set of points $\mathcal{X}$ describing the selected component. The oriented bounding box is then defined by
\begin{mydefi}
Given a finite  set  of $N$ points $\mathcal{X} = \{X_i \mid i = 1, ... , N\} \subset \mathbb{R}^3 $, the optimal \emph{oriented bounding box} (OBB) is defined as the arbitrarily oriented cuboid, or rectangular  parallelepiped, of minimal volume enclosing $\mathcal{X}$.
\end{mydefi}
Each OBB is uniquely defined by its center $\mathbf{X} \in \mathbb{R}^3$, its rotation $R \in SO(3,\mathbb{R})$, and its extensions $\mathbf{\Delta}  \in \mathbb{R}^3$, where $SO(3,\mathbb{R}) = \{R \in \mathbb{R}^{3 \times 3} \mid R^T R = I_d = RR^T, det(R) = 1\}$ are all the orthogonal and real $3$-by-$3$-matrices. Given the definition, the optimal oriented bounding box is 
the solution to a constrained optimization problem
\begin{equation} \label{def_con_op}
\begin{matrix} 
\underset{ 
\begin{matrix}
\mathbf{\Delta},\mathbf{X} \in \mathbb{R}^3 \\
R \in SO(3,\mathbb{R})
\end{matrix}
}{\text{min}
} & 
\prod_{k=1}^d \mathbf{\Delta}_k \\

\text{s.t.} &
-\frac{1}{2}\mathbf{\Delta} \leq R X_i - \mathbf{X} \leq \frac{1}{2} \mathbf{\Delta} &
 \forall i = 1,\dots,N\\

\end{matrix}.
\end{equation}
The matrix $R$ rotates the standard coordinate system of $\mathbb{R}^3$, such that the bounding box is axis aligned in the rotated reference frame.

We employ the HYBBRID algorithm \cite{ChangBox} to approximate the boxes, since the exact algorithm has cubic runtime in the number of points in the convex hull of $\mathcal{X}$ \cite{RourkeBox}. Also, the quality of the approximation by fast principal component analysis (PCA) based algorithms is limited \cite{DimiPCABox}. However, we need the oriented bounding boxes to fit tightly, which is why we adapted the HYBBRID algorithm for the OBB approximations of time sequences.

\subsubsection*{Adapted HYBBRID Algorithm} A redefinition of (\ref{def_con_op}) as in \cite{ChangBox} allows the problem's analysis as an unconstrained optimization over $SO(3,\mathbb{R})$
\[  \underset{R \in SO(3,\mathbb{R})}{\text{min}} f(R) , \]
where $f(R)$ is the volume of the axis aligned bounding box of the point set $\mathcal{X}$ rotated by $R$. 
Since the function is not differentiable, a derivative-free algorithm combining global search by a genetic algorithm \cite{Goldberg89,Holland75} (exploration) and local search by the Nelder-Mead simplex algorithm \cite{NelderMead} (exploitation) to find the rotation matrix $R^* \in SO(3,\mathbb{R})$ that minimizes $f$ is introduced in \cite{ChangBox}, where a detailed description of the algorithm can be found.

By utilizing the best rotation matrix from the previous time step of the same component, we can give a good initial guess to the algorithm. This improves the quality of the approximations and reduces the runtime by 30\% in comparison to the original HYBBRID algorithm.

\subsubsection*{Information Retrieval from OBBs}
The oriented bounding boxes further allow consideration of translation, rotation, and scale of the car components individually.
The rigid transform of a component, composed of its translation and rotation, is normally estimated using a few fixed reference points for each component \cite{rigidbody} and might be subtracted from the total deformation for analysis. Nevertheless, the quality of this rigid transform estimation depends highly on the hand selected reference points. 
When utilizing oriented bounding boxes the use of reference points is not necessary any more, since the HYBBRID algorithm outputs the rotation matrix $R^*$. This allows detecting interesting characteristics in the components’ deformations that are not translation or rotation based.

\section{LSTM Autoencoders} 
\label{sec:LSTM}

Autoencoders are a widely used type of neural network for unsupervised learning. They learn low dimensional characteristic features of the input that allow its recreation as well as possible. Therefore, the hidden or low dimensional re\-pre\-sen\-ta\-tions should show the most relevant aspects of the data and classification tasks can generally be solved better in the low dimensional space. Autoencoders stand out against other dimension reduction approaches, because the encoder enables quick evaluation and description of the manifold, on which the hidden representations lie \cite{Goodfellow}. Therefore, a new data point’s embedding does not have to be obtained by interpolation.

Since in the case of simulation data we are always handling time sequences, we implemented an autoencoder as in \cite{UnsupLSTM}, using long short-term memory (LSTM) layers \cite{Graves13,Hochreiter} in the encoder and decoder. LSTM are a type of recurrent neural networks (RNN) \cite{Rumelhart} that apply parameter sharing over time and thus reduce the number of trainable weights. By this means, we can analyze the deformation process of a structural component over time, because the information from all the input time steps is summarized into one hidden representation taking advantage of the temporal correspondences in the data.

\begin{figure}[t]
\center
\begin{tikzpicture}[scale=0.4, every node/.style={scale=0.85}]

\fill [gray, rounded corners, opacity=0.15] (-2, -1.3) rectangle (16, 1.5);
\node[align=center,gray] at (13,0.1) {ENCODER};

\draw[thick,->] (-1,1.1) to[out=-90, in=180] (0,0.1);
\draw[rounded corners] (0, -1) rectangle (7, 1.2) {};
\node[align=center] at (3.5,0.1) { LSTM layer };

\draw[thick] (-1,1.1) -- (-1,1.9);
\node[align=center] at (-1,2.4) {$S_{in} =  \{s_1, s_2, \dots, s_{T_{IN}}\}$};

\draw[thick] (7,-0.4) --  (7+1,-0.4);
\draw[thick,->] (7+1,-0.4) to[out=0, in=90]  (7+2,-2.5);
\node[align=center] at (10,-1.6) {$h_{T_{IN}}$};

\node[align=center, ] at (7.3,-3.3) {\textbf{low-dim representation:} $ h_{T_{IN}}$};
\draw[thick, ->] (1.7,-3.4) to[out=180, in=90] (0.8,-4.9);
\draw[thick, ->] (12.9,-3.4) to[out=0, in=90] (13.8,-4.9);

\fill [gray, rounded corners, opacity=0.15] (-7, -4.6) rectangle (6.9, -10.9);
\node[align=center,gray] at (-4,-6.1) {1\textsuperscript{st} DECODER};
\draw[rounded corners] (-0.5, -4.9) rectangle (6.3, -7.3) {};
\node[align=center] at (3,-6.1) { LSTM layer }; 

\fill [gray, rounded corners, opacity=0.15] (7.7, -4.6) rectangle (21.6, -10.9);
\node[align=center,gray] at (18.6,-6.1) {2\textsuperscript{nd} DECODER};
\draw[rounded corners] (8.3, -4.9) rectangle (15.1, -7.3) {};
\node[align=center] at (11.7,-6.1) { LSTM layer }; 

\draw[thick, ->] (0.8,-7.3) -- node[left] {$\{h'_1, h'_2, \dots, h'_{T_{IN}}\}$} (0.8,-9);
\draw[rounded corners] (0.8-3.1, -9) rectangle (0.8+5.5, -9-1.6) {};
\node[align=center] at (0.8+1.4,-9.8) { Feed forward layer $\textbf{F}_1$ };

\draw[thick, ->] (13.8,-7.3) -- node[right] {$\{h'_{T_{IN}+1}, \dots, h'_{T_{FIN}}\}$} (13.8,-9);
\draw[rounded corners] (13.8+3.1, -9) rectangle (13.8-5.5, -9-1.6) {};
\node[align=center] at (13.8-1.3,-9.8) { Feed forward layer $\textbf{F}_2$ };

\draw[thick, ->] (0.8,-10.6) -- (0.8,-10.6-1.2);
\node[align=center] at (0.8,-10.6-1.8)  {$S_{reconstruct} =  \{o_1, o_2, \dots, o_{T_{IN}}\}$} ;

\draw[thick, ->] (13.8,-10.6) -- (13.8,-10.6-1.2);
\node[align=center] at (13.8,-10.6-1.8)  {$S_{predict} =  \{o_{T_{IN}+1}, \dots, o_{T_{FIN}}\}$} ;

\end{tikzpicture}
\caption[LSTM autoencoder]{LSTM autoencoder. The last hidden vector $h_{T_{in}}$ of the encoder layer is utilized as the low dimensional representation, which is input to all the LSTM units of the decoders. By applying time independent feed forward layers to the hidden vectors $h'_\bullet$ of each decoder, their size is reduced and the output sequences $S_{reconstruct}$ and $S_{predict}$ are created. The weights of the LSTM layers and $\textbf{F}_\bullet$ are time independent. }
\label{lstm_auto}
\end{figure}
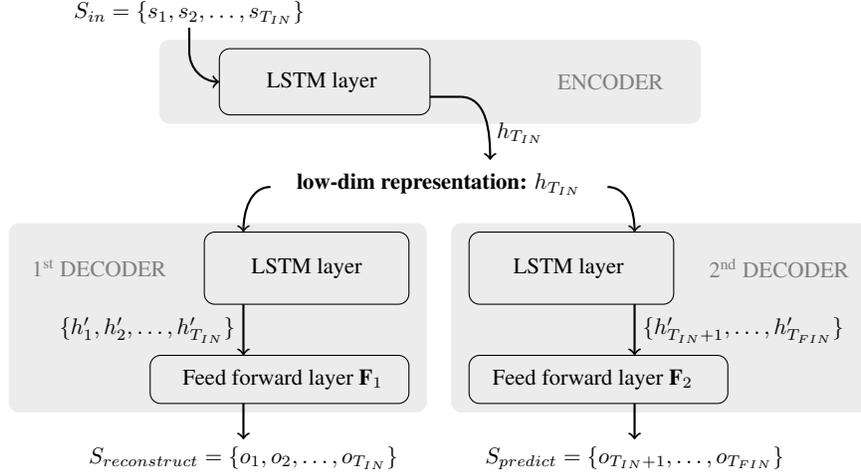

Because we want to analyze the time sequences in-situ, we not only recreate the input, but also predict the future time steps by implementing a second decoder \cite{UnsupLSTM}, as illustrated in figure \ref{lstm_auto}. 
The prediction gives a first estimate of the simulation result.
Furthermore, the low dimensional embedding is expected to be more significant with respect to the detection of deformation modes in future time steps, since the low dimensional features are also trained to predict future deformation.

We take advantage of the good generalization performance of autoencoders and train an LSTM autoencoder that can handle all the differently shaped components and generalizes their deformation behavior represented by oriented bounding boxes. 

Therefore, we separately feed the oriented bounding boxes of each structural component for the first $T_{IN}$ time steps to the autoencoder.
As outputs, the autoencoder produces a hidden representation $h_{T_{IN}}$ and a reconstruction of the input sequence $S_{reconstruct}$. Since we implement two decoders, we also obtain a prediction $S_{predict}$, as illustrated in figure \ref{lstm_auto}.
We chose to use only one LSTM layer for each en-/decoder, to minimize the number of trainable weights and impede overfitting. Depending on the size of the training set, an increase in the number of layers might improve the prediction and representation results.

\section{Related Work} 
\label{sec:rel}

The analysis of car crash simulations has been studied with different machine learning techniques. For example, 
\cite{bohn2016sparse,bohnTRUCK,shapemining14,Zhao10} 
base their analysis on the finished simulation runs, whereas we focus on in-situ analysis of car crash simulations.

Other recent works study analysis and prediction of car crash simulation data with machine learning under different problem definitions than in this work, for example the estimation of the result given the input parameters \cite{Guennec2018} or plausibility checks  \cite{MechDL}. For a recent overview on the combination of machine learning and simulation see \cite{Rueden.ea:2020}.

There are alternatives to oriented bounding boxes as low dimensional shape representations for 3D objects. The authors of \cite{iza2019geometrical} construct low dimensional shape representations via a projection of mesh data onto an eigenbasis stemming from a discrete Laplace Beltrami Operator. 
The data is then represented by the spectral coefficients. We have chosen a simpler representation, that can also be applied to point clouds and allows a faster training.

The majority of works about neural networks for geometries extend convolutional neural networks on non-euclidean shapes, including graphs and manifolds.
The works in \cite{BronsteinGDL,litany2017,MontiMixture17} present extensions of neural networks via the computation of local patches.
Nevertheless, the networks are applied to the 3D meshes directly and are computationally expensive, when applied for every time step and component of complex models. 

Based on the shape deformation representation by \cite{gao_meshdef}, recent works studied generative modeling of meshes for 3D animation sequences via bidirectional LSTM network \cite{bilstm_mesh} or variational autoencoder \cite{tan_var_autoencoderA}.
The shape representation yields good results, however, it solves an optimization problem at each vertex and requires identical connectivity over time.
The architectures are tested on models of humans with considerably fewer nodes, which is why computational issues seem likely for the vertex-wise optimization problem.  

\section{Experiments} 
\label{sec:exp}

\begin{figure}[t]
\center
\begin{minipage}[c]{\textwidth}
\center
\includegraphics[width=0.45\textwidth,trim={8cm 2.5cm 12cm 3.4cm},clip]{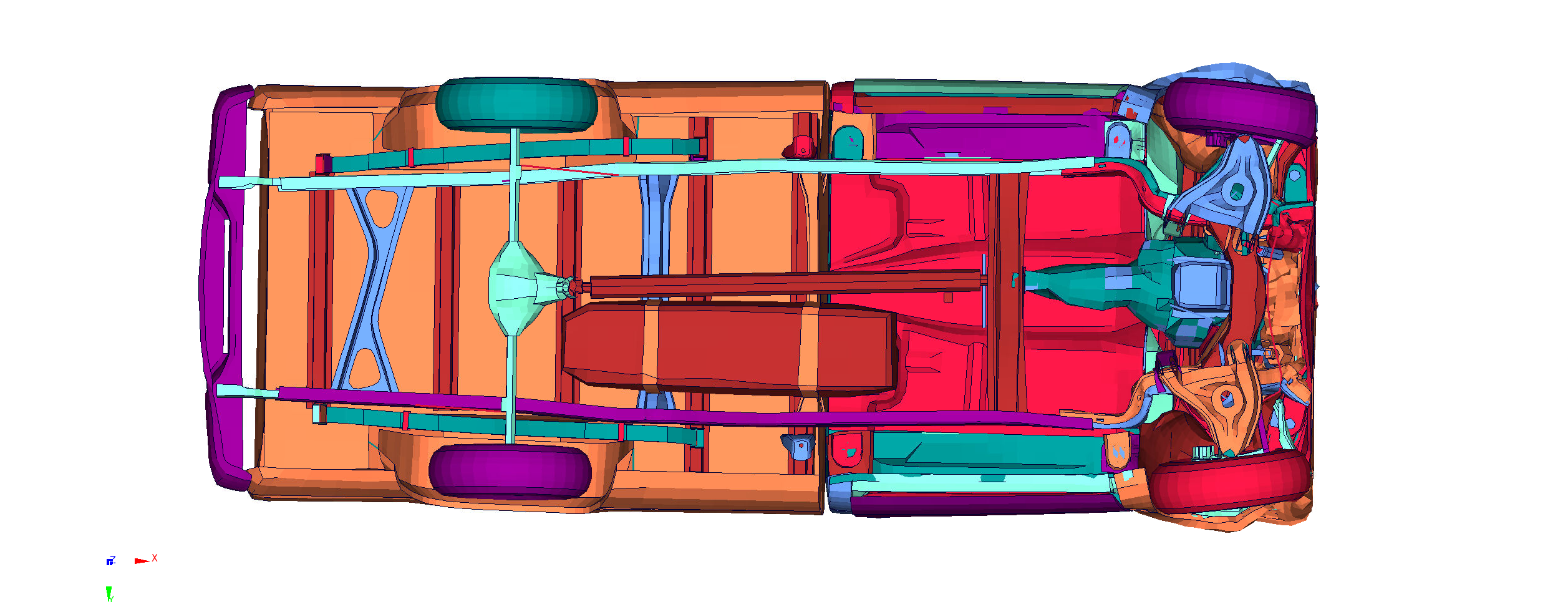}
\includegraphics[width=0.45\textwidth,trim={10cm 6cm 7cm 3.5cm},clip]{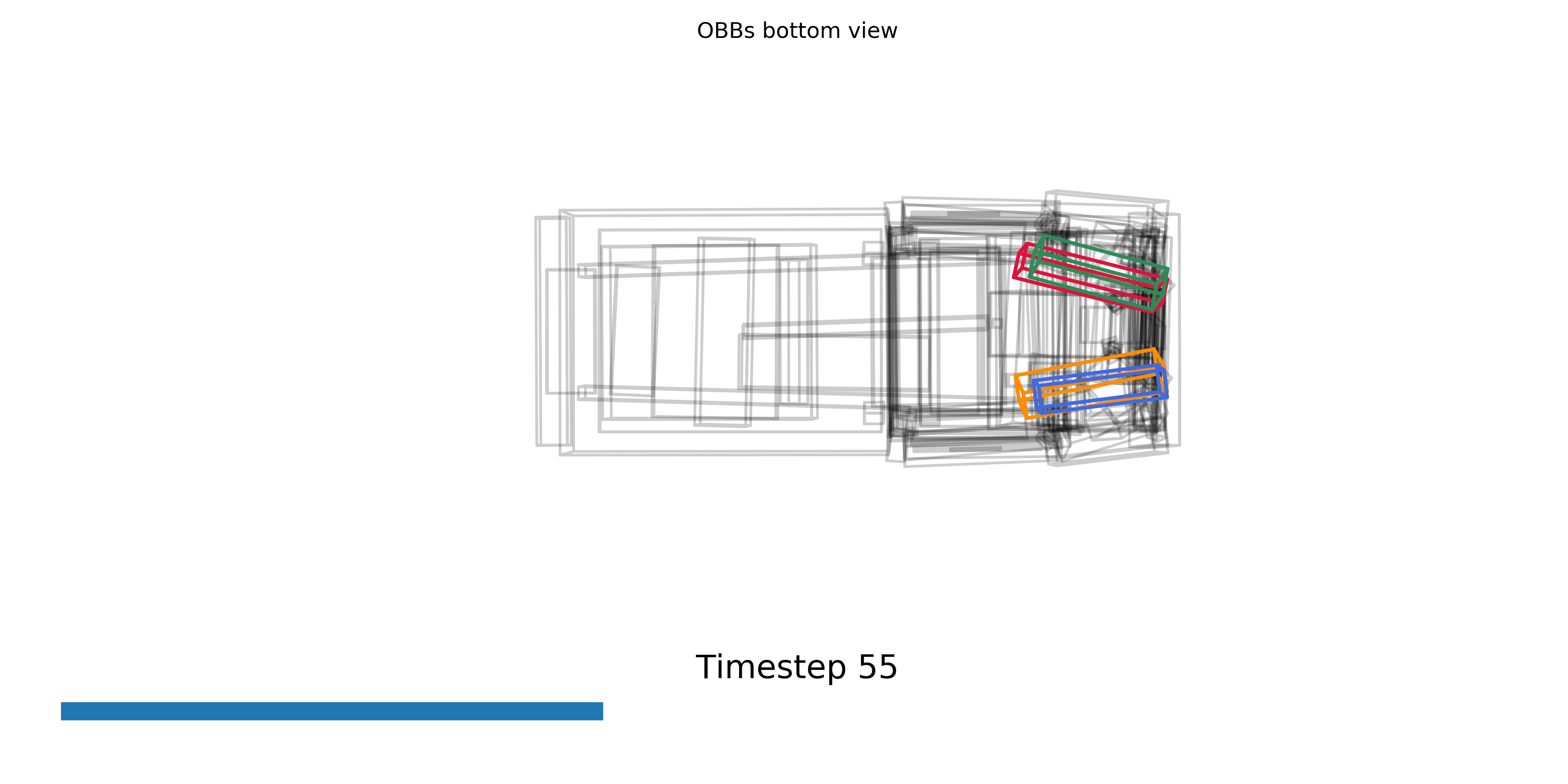}
\caption{Snapshot and bounding box representation of the example car model at $T_{IN}=12$ (bottom view). The four beams from figure \ref{method} are colored.} \label{snapshot1}
\end{minipage}
\end{figure}

We evaluated the introduced approach on a frontal crash simulation of a Chevrolet C2500 pick-up truck\footnote{from NCAC \url{http://web.archive.org/web/*/www.ncac.gwu.edu/vml/models.html}}.
The data set consists of $n_{simulations} = 196 $ completed simulations\footnote{computed with LS-DYNa \url{http://www.lstc.com/products/ls-dyna}} of a front crash (see figure \ref{snapshot1}), using the same truck, but with different material characteristics, which is a similar setup to \cite{bohnTRUCK}. For 9 components (the front and side beams and the bumper parts) the sheet thickness has been varied while keeping the geometry unchanged.
From every simulation we select $T_{FIN} = 31$ equally distributed time steps and use $T_{IN} = 12$ time steps for the input to the in-situ analysis, which is applied to 133 structural components represented by oriented bounding boxes. That means, when detecting a bad simulation run after $T_{IN}$ of $T_{FIN}$ time steps the design engineer can save 60\% of the simulation time.
The impact at $T_{IN} = 12$ is visualized in figure \ref{snapshot1}. At this time most of the deformation and energy absorption took place in the so-called crumble zone at the front of the car.
\begin{table}[t]
\center
\small
\caption{Structure of the composite LSTM autoencoder with two decoders for reconstruction and prediction. The bullets ${\scriptstyle \bullet}$ reference the corresponding batch size.} \label{network_composite}
\begin{tabular}{c|ccc}
Layer & Output Shape & Param. & Connected to \\ \hline
Input & $({\scriptstyle \bullet},T_{IN},24)$ & 0 \\
LSTM 1 & $({\scriptstyle \bullet},24)$ & 4704 & Input \\
Repeat Vector 1 & $({\scriptstyle \bullet},T_{IN},24)$ & 0 & LSTM 1\\
Repeat Vector 2 & $({\scriptstyle \bullet},T_{FIN} - T_{IN},24)$ & 0 & LSTM 1\\
LSTM 2 & $({\scriptstyle \bullet},T_{IN}, 256)$ & 287744 & Repeat Vector 1 \\
LSTM 3 & $({\scriptstyle \bullet},T_{FIN} - T_{IN}, 256)$ & 287744 & Repeat Vector 2 \\
Fully Connected 1 & $({\scriptstyle \bullet},T_{IN},24)$ & 6168 & LSTM 2\\
Fully Connected 2 & $({\scriptstyle \bullet},T_{FIN} - T_{IN},24)$ & 6168 & LSTM 3\\
\end{tabular}
\end{table}

The data is normalized to zero mean for each of the three dimensions and standard deviation one over all the dimensions.
Using all 133 components from 100 training samples we train the network\footnote{implemented in Keras \cite{chollet2015keras}, no peephole connections} for 150 epochs with adaptive learning rate optimization algorithm \cite{adamopt}.  We minimize the mean squared error between true and estimated bounding boxes over time, summing up the error for reconstruction and prediction decoder as well as for the components. The remaining 96 simulations are testing samples.

Because of the limited number of training samples, we tried to minimize the number of hidden neurons while maintaining the quality, to reduce the possibility of overfitting and speed up the training. We observed that the result is stable with respect to the number of hidden neurons, which we finally set to 24 for the encoder and 256 for the decoders. This leads to a total of 592,528 trainable weights, which are distributed over the layers as listed in table \ref{network_composite}. The training has a runtime of 36 seconds for one epoch on a CPU with 16 cores.

\subsection{Prediction of Components' Deformations during a Crash Test}

The predicted oriented bounding boxes give an estimate after $T_{IN}$ time steps for the result of simulation $S$ afterwards. 
For comparison, we define two baselines using a nearest neighbor search, where the prediction of $S$ is estimated by the future time steps of another simulation $S'$, which is either the corresponding simulation for the nearest neighbor of $S$ in the input sequences or in the input parameter space. We compare the results by their mean squared error which is also used for training. 
Note that an interpolation has not been chosen for comparison, because interpolated rectangular boxes can have any possible shape.

The prediction error of our method is 38\% lower than a prediction based on nearest neighbor search in the input parameter space and 16\% lower than a prediction using the nearest neighbor of the training input sequences, see table \ref{mse}. 
We also compare the LSTM composite autoencoder to a simple RNN composite autoencoder, to evaluate the use of the LSTM layers in comparison to a different network architecture. When using the same number of hidden weights for the simple RNN layer in the encoder and the decoders, the prediction error is higher than for the LSTM Autoencoder. This indicates, that the LSTM layers better detect the temporal correspondence in the time sequences.
Additionally, the results depending on random initial weights are stable, which is demonstrated by the low standard deviation of the error.

\begin{table}[t]
\small
\center
\caption{Mean Squared Errors ($\times 10^{-4}$) on testing samples for different prediction methods applied to oriented bounding boxes. 
For methods depending on random initialization, mean and standard deviation of 5 training runs are listed.
}\label{mse}
\begin{tabular}{l|l|c}
Method &  \hspace{2ex}Test-MSE\hspace{2ex}  & \hspace{1ex}STD\hspace{1ex}\\
\hline
Nearest neighbor in parameter space &  \hspace{4ex}13.34 & - \\
Nearest neighbor in input sequences & \hspace{5ex}6.13  & -\\ 
Composite RNN autoencoder &  \hspace{5ex}5.80 & \hspace{1ex} 0.095 \hspace{0.5ex} \\
Composite LSTM autoencoder &  \hspace{5ex}5.14 & \hspace{1ex} 0.127 \hspace{0.5ex} \\
\end{tabular}
\end{table}

We observe an improvement in the orthogonality between the faces of the rectangular boxes over time for both the prediction and reconstruction output of the LSTM layers. The LSTM layers recognize the conditions on the rectangular shape and implicitly enforce it in every output time step. Note, the network generalizes well to the highly different sizes and shapes of 133 different structural components, whose oriented bounding boxes are illustrated in figure \ref{snapshot1}.

\subsection{Clustering of the Deformation Behavior}

The LSTM autoencoder does not only output a prediction, but also a hidden low dimensional representation. It captures the most prominent and distinguishing features of the data, because only based on those the prediction and reconstruction are calculated. Its size corresponds to the number of hidden neurons in the encoder. Hence, our network reduces the size of the $24\times T_{IN}$-dimensional input to $24$. The hidden representation summarizes the deformation over time, which is why we obtain one hidden representation for the whole sequence of each structural component and simulation.

In the case of car crash simulations we are especially interested in patterns in the plastic deformation to detect plastic strain and material failure. Therefore, we add an additional preprocessing step for the analysis of deformation behavior. We subtract the rigid movement of the oriented bounding boxes, that means we translate the oriented bounding boxes to the origin and rotate them to the standard coordinate system (see section \ref{sec:obb}). In that way only the plastic deformation and strain are studied, but not induced effects in the rigid transform.

For a subset of components we visualize the 24-dimensional hidden representations in two dimensions with the t-SNE dimension reduction algorithm \cite{tsne}. Figure \ref{method} illustrates the embedding for the four frontal beams. 
We notice that the deformation mode from figure \ref{snapshot2} is clearly manifesting in the hidden representations of the left frontal beams.

\begin{figure}
\center
\begin{minipage}[c]{\textwidth}
\center
\includegraphics[width=0.4\textwidth,trim={6.8cm 4.5cm 13.3cm 4.5cm},clip]{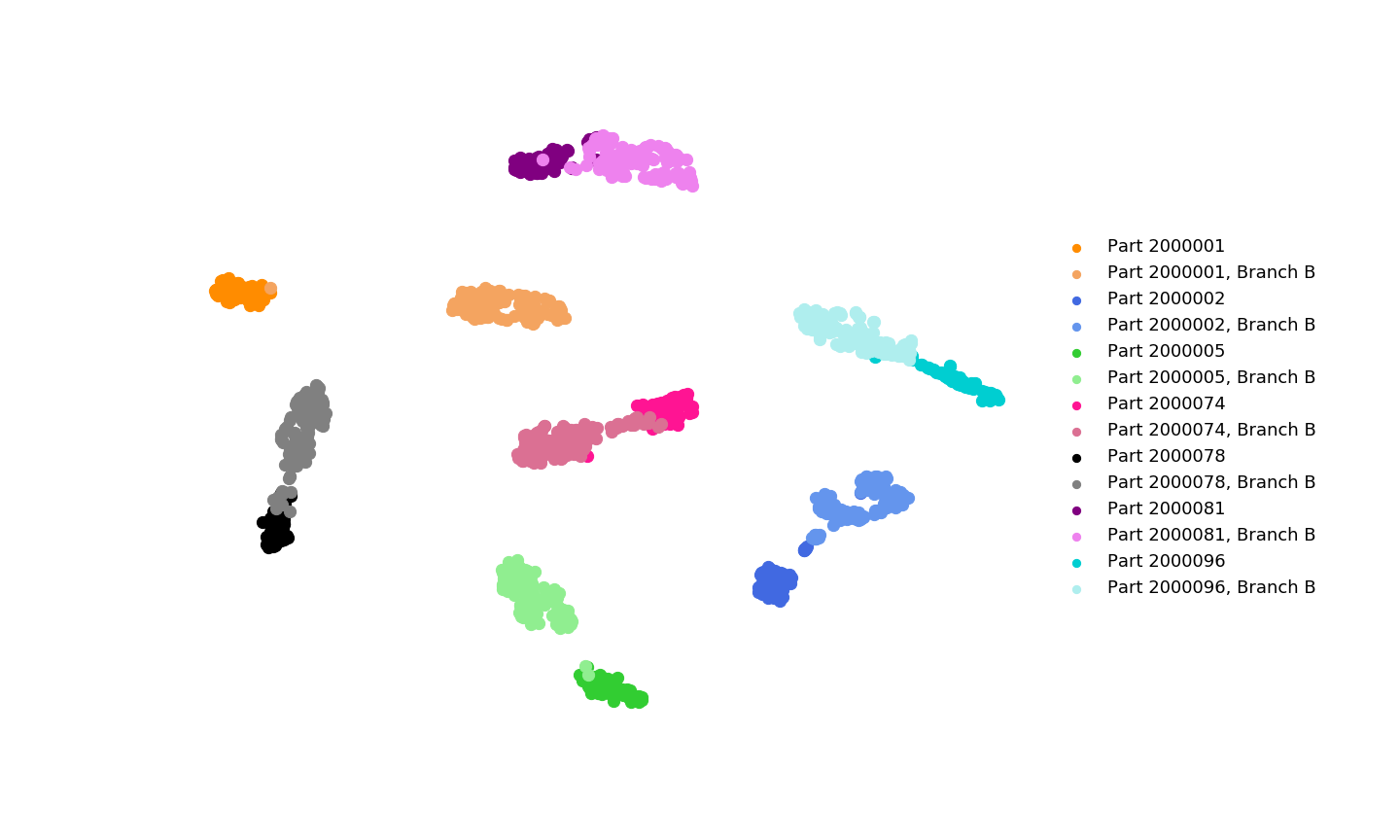}
\includegraphics[width=0.54\textwidth, trim={30.5cm 6.5cm 4.8cm 5.9cm},clip]{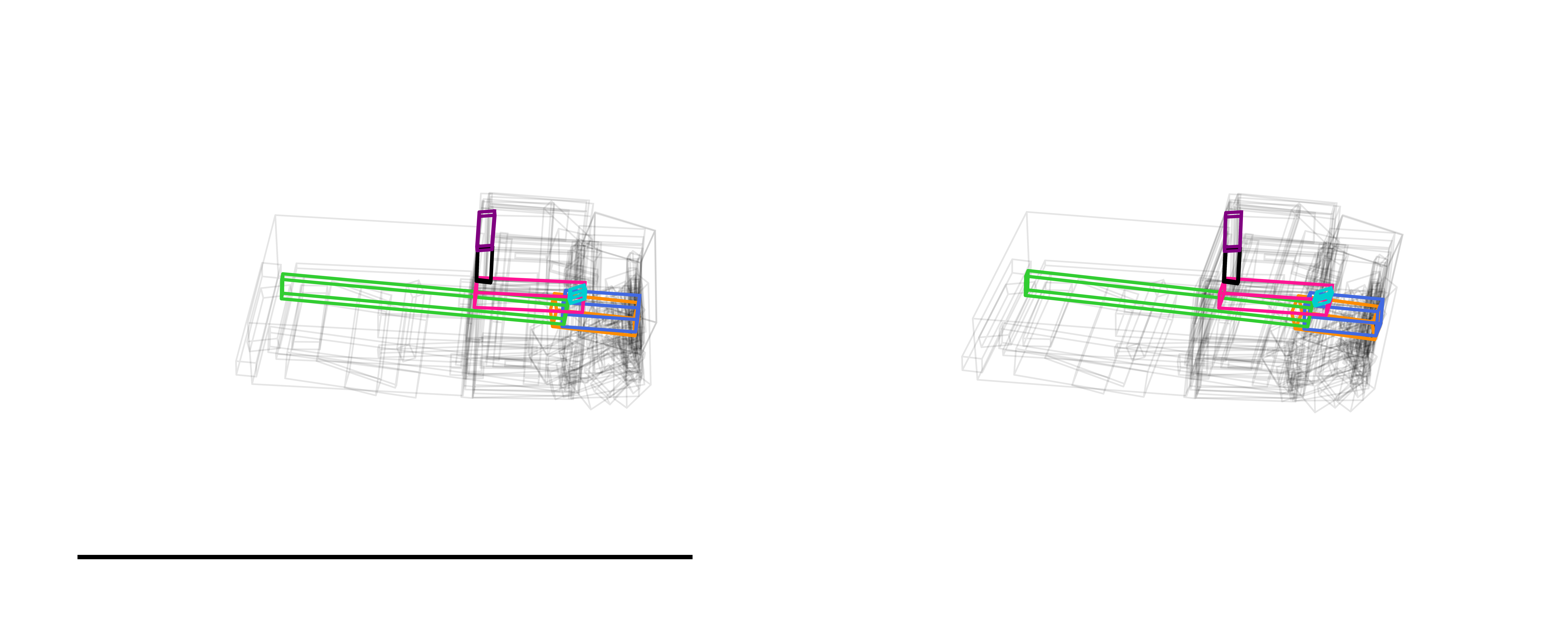}
\caption{Embedded hidden representations for all simulations and seven structural components, that manifest the deformation modes from figure \ref{snapshot2}. The two different patterns for each of the selected structural components are illustrated in the same color with different intensity, the corresponding component on the right in the matching color.} \label{defmode}
\end{minipage}
\end{figure}

When observing the hidden representations of other components, we can detect the pattern corresponding to the deformation mode from figure \ref{snapshot2} in other structural components, whose plastic deformation is seemingly influenced by the behavior of the left frontal beams.  Even though the interaction between structural components is not fed to the network, the components that are affected by the mode can be identified in figure \ref{defmode}. 

\section{Conclusion}

We have presented a general approach for in-situ analysis and prediction of deforming 3D shapes by subdividing them into structural components. The simple shape representation makes an analysis of complex models feasible and allows the detection of coarse deformation patterns.
The method is applied successfully to a car crash simulation, estimates the final position of the boxes to a higher quality than other methods, and speeds up the selection of new simulation parameters without having to wait for the final results of large simulations.

Although we have selected a relatively simple shape representation, the patterns can be reliably detected. In future work, we plan to compare and combine the approach with other low dimensional shape representations and investigate other application scenarios, in particular in engineering.
Apart from simulation data, our approach has more application areas, including the analysis and prediction of the movement of deformable articulated shapes, to which  \cite{articulated} gives an overview. The authors present data sets for human motion and pose analysis, on which the introduced approach can also be applied.


\bibliographystyle{splncs04}
\bibliography{mybibliography}

\end{document}